\documentclass[letterpaper, 10 pt, conference]{ieeeconf}  % Comment this line out if you need a4paper

\IEEEoverridecommandlockouts                              % This command is only needed if 
\usepackage{subcaption}
\usepackage{dblfloatfix}
\usepackage{float}
\usepackage{graphicx}
\let\labelindent\relax
\usepackage{enumitem}
\graphicspath{ {./images/} }

\title{\LARGE \bf
Wireless Network Demands of Data Products from Small Uncrewed Aerial Systems at Hurricane Ian
}

\author{Thomas Manzini$^{1}$, Robin Murphy$^{1}$, David Merrick$^{2}$, and Justin Adams$^{2}$% <-this % stops a space
\thanks{$^{1}$ Department of Computer Science and Engineering,
        Texas A\&M University,
        College Station, TX 77843, USA
        {\tt\small tmanzini@tamu.edu, robin.r.murphy@tamu.edu}}%
\thanks{$^{2}$Center for Disaster Risk Policy at Florida State University,  Tallahassee, FL 32306, USA
        {\tt\small dmerrick@em.fsu.edu}, {\tt\small jadams@em.fsu.edu}}%
}

\begin{document}

\maketitle
\thispagestyle{empty}
\pagestyle{empty}

%%%%%%%%%%%%%%%%%%%%%%%%%%%%%%%%%%%%%%%%%%%%%%%%%%%%%%%%%%%%%%%%%%%%%%%%%%%%%%%%
\begin{abstract}
Data collected at Hurricane Ian (2022) quantifies the demands that small uncrewed aerial systems (UAS), or drones, place on the network communication infrastructure and identifies gaps in the field. 
Drones have been increasingly used since Hurricane Katrina (2005) for disaster response, however getting the data from the drone to the appropriate decision makers throughout incident command in a timely fashion has been problematic. These delays have persisted even as countries such as the USA have made significant investments in wireless infrastructure, rapidly deployable nodes, and an increase in commercial satellite solutions. 
Hurricane Ian serves as a case study of the mismatch between communications needs and capabilities. In the first four days of the response, nine drone teams flew 34 missions under the direction of the State of Florida FL-UAS1, generating 636GB of data. The teams had access to six different wireless communications networks but had to resort to physically transferring data to the nearest intact emergency operations center in order to make the data available to the relevant agencies. 
The analysis of the mismatch contributes a model of the drone data-to-decision workflow in a disaster and quantifies wireless network communication requirements throughout the workflow in five factors. Four of the factors-availability, bandwidth, burstiness, and spatial distribution-were previously identified from analyses of Hurricanes Harvey (2017) and Michael (2018). This work adds upload rate as a fifth attribute. The analysis is expected to improve drone design and edge computing schemes as well as inform wireless communication research and development.

%% TOM: use Heilmeier's catechism as per the How to Write Papers in the denizen folder
% During the response to Hurricane Ian FL-UAS1, the robotic and remote sensing group, encountered operational limitations as a result of the wireless infrastructure that was stood up during the response. 
% These limitations hampered the robotic response and caused significant delays between when data was collected in the field and when it could be presented to decision makers. 
% This paper details the data flow associated with running missions in the Fort Myers area during the response to Hurricane Ian. 
% It motivates the need for improvements to wireless infrastructure built in response to a large scale disaster by showing the disparities between operational needs and current capabilities. 
% And finally, it concludes by detailing future directions for research and operations with respect to these kinds of large scale disasters.

\end{abstract}

%%%%%%%%%%%%%%%%%%%%%%%%%%%%%%%%%%%%%%%%%%%%%%%%%%%%%%%%%%%%%%%%%%%%%%%%%%%%%%%%
\section{INTRODUCTION}

%% TOM: Add the map. But the red oval is too small and does not show wauchula
\begin{figure}
    \centering
    \includegraphics[width=8.5cm]{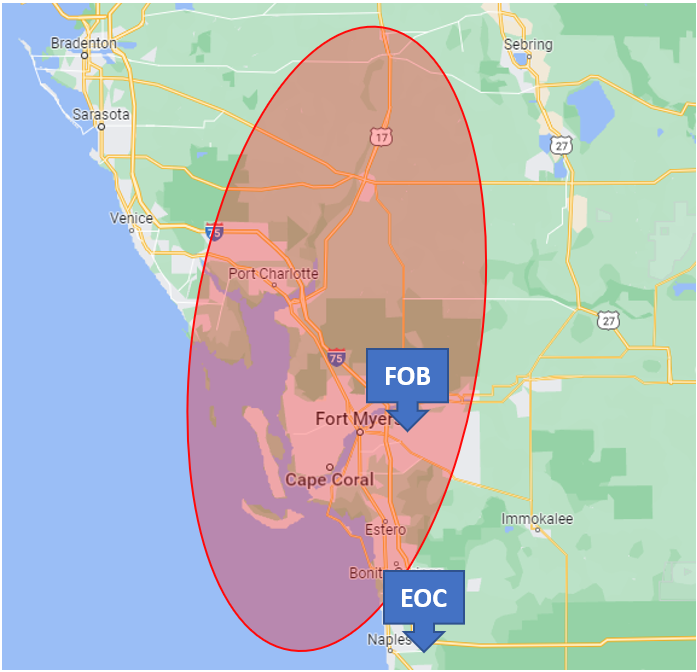}
    \caption{The Hurricane Ian drone area of operations. The general region of where missions were located is marked in the red oval. The Forward Operating Base (FOB) and Emergency Operations Center (EOC) are marked in blue.}
    \label{fig:map}
\end{figure}

Hurricane Ian was one of the most destructive hurricanes to hit the state of Florida, resulting in 146 fatalities and \$40B of property damage \cite{wiki:Ian}. From September 27 to October 17, 2022,  the Florida State Emergency Response Team  FL-UAS1 task force, led by Florida State University's (FSU) Center for Disaster Risk Policy, was charged with conducting drone missions to meet requests of agencies having jurisdiction.  
FL-UAS1 coordinated teams and response experts from 
nine agencies (Alachua County Fire Rescue, Boone County Fire Protection District (Missouri), Florida Department of Law Enforcement, Jacksonville Sheriff's Office, Leon County Sheriff's Office, Miami-Dade Fire Rescue, Okaloosa County Sheriff's Office, Tallahassee Fire Department, Tallahassee Police Department), 
two institutions (FSU, Texas A\&M), and 
one insurance company (USAA).  

 Fig.~\ref{fig:map} shows the general area of coverage with the Forward Operating Base (FOB) established on Sept 30.  It was the fourth time where ten or more drone squads were formally used as part of a consolidated state or local effort in the USA (others were Hurricanes Harvey (2017), Michael (2018), and Florence (2018)). The Ian missions covered more area than previous hurricanes, mapping the entirety of Fort Myers Beach, Naples, Pine Island, Sanibel Island, Boca Raton, other  surrounding islands and impacted residential areas of Lee and Charlotte Counties, and inland through Wauchula.

Understanding the missions, operational tempo, and conditions for the first four days (Sept. 29-Oct. 2) of the Ian effort is important in establishing the wireless data demands and gaps as it represents the immediate life saving response and mitigation phase. During that phase, FL-UAS1 teams conducted 34 missions and collected more than 636GB of data. Although the teams brought six different wireless hotspot devices (Starlink: RV Package, Business Package; FirstNet: Sonim, custom PSEC router, and personal phones: AT\&T, Verizon), the amount of data exceeded the capability and the team had to resort to ``war driving" to Florida Gulf Coast University 40 minutes away. That wireless network did not have sufficient capacity and so data managers made nightly trips to Collier County Emergency Operations Center, the nearest intact agency with wired infrastructure. 

The Hurricane Ian deployment provides a unique opportunity to 
quantify for the first time the wireless network communication capabilities needed to deliver data products from drones to agencies. In particular, it establishes how network availability and demands evolve over time (\textit{availability}), their \textit{bandwidth}, the occurrence of \textit{burstiness} in transmitting data, the \textit{spatial distribution} of members of the drone data-to-decision process, and the \textit{upload rates}. 
Understanding the demands on the wireless infrastructure after a disaster and how it differs from normal conditions is expected to enable both roboticists to design more effective systems and the wireless communications community to better support higher throughput and Cloud computing capabilities. The analysis is limited, as observing the condition of the wireless networks at Ian was opportunistic and secondary to primary goals of providing data products. It should be noted that the  observations focused on the volume, types, rates, and spatial distribution, not network metrics such as received signal strength indicator \cite{channelMeasurements}. 
%Channel Measurements and Modeling in Drone-to-Ground Communications- 

\section{RELATED AND PREVIOUS WORK}

%%%: TOM: NO. DEFER THIS LEVEL OF MOVING ABOUT UNTIL ALL THE TEXT AND FIGURES ARE IN.  Moving this to the related work section so it appears on the same page as the data to decision workflow section. 

This paper differs from related work in wireless communications for drones in that it focuses on throughput throughout the data-to-decision process, not just the to the operational team or for Cloud post-processing of data.
%operational team, data managers, Cloud post-processing, distribution to end decision makers). 
It quantifies previous work in emergency informatics by empirically capturing the inverse relationship between the availability of actionable data during the response phase with the impact of that data \cite{ccc:EmergencyInformatics}.  Hurricane Ian results are consistent with Hurricane Harvey \cite{ssrr18:harvey} and Michael \cite{SSRR19:michael}, confirming the four factors on wireless networks and adding a fifth, upload rate. 

Generally, the robotics literature focuses on i) use of drones to establish wireless networks to support general communications or communicate with the operational team on the ground, sometimes collectively referred to as the ``Internet of Drones" \cite{InternetOfDrones},   or ii) applications  of wireless enabled drones \cite{WSN}. Similar topics that are also out of the scope of this paper include: 
using drones to establish comms or establishing specialized networks for drones \cite{commArchitecture,
flightSchemes}, how to optimize layers in the comms stack to accommodate throughput bottlenecks in drones \cite{WMN}, identification of 
better frequencies  or mechanisms for drone to ground comms \cite{channelMeasurements},
and off-loading or selection of network resources \cite{DroneCOCoNet}. 

\begin{figure}[t]
    \centering
    \includegraphics[width=9cm]{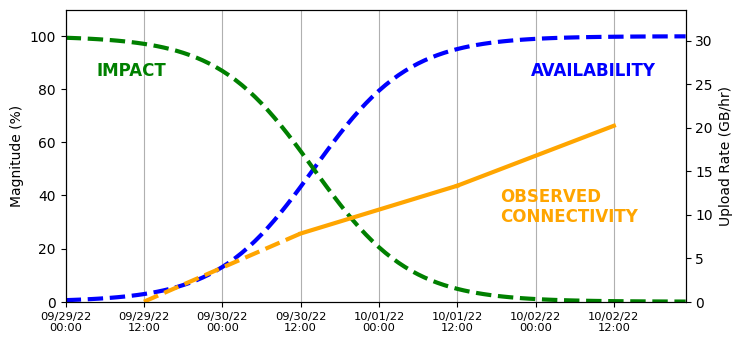}
    \caption{Impact (green dashed) versus availability of information (blue dashed) to emergency responders, adapted from \cite{ccc:EmergencyInformatics}. Scaled availability of drone data products as \% update rate over wireless networks at Hurricane Ian in gold.}
    \label{fig:informationCurve}
\end{figure}

A 2010 white paper prepared for the Computing Community Consortium on gaps in the general data-to-decision workflow during a disaster described a misalignment shown in Fig.~\ref{fig:informationCurve} between the availability of information for response (blue) and the impact (green). The misalignment could be mitigated with i) new methods of data acquisition, such as drones, and ii) real-time transmission of the data via faster, higher capacity networks. This paper addresses both mitigations. Actual deployments of drones at disasters by the authors \cite{ssrr18:harvey,SSRR19:michael,murphy:computer2016} suggest that the real-time transmission of drone data can be characterized by four factors: \textit{availability} of communications  over the first 72 hours (the nominal duration of the response phase), the \textit{bandwidth} capability of the communications, the \textit{burstiness} of data, that is, how much of the data should be transmitted in batches rather than continuous streaming, and the \textit{spatial distribution} of access to the network. Hurricane Ian extends these factors by being the first to document wireless upload and download rates.

% Availability during disaster
% CCC paper
% our papers: Harvey, Michael
% No formal model or prediction of restoration, typically assume the 72 hours (IEEE Xplore, Science Direct)
% Bandwidth demand
% Harvey, Michael described volume by day but not rates
% No discussion of bandwidth (IEEE Xplore, Science Direct)
% Bursty
% 2015 Computer.org paper
% Spatial distributed data-to-decision 
% mine and jenny’s paper
% Interference among users leading to marketplace: https://doi.org/10.1016/j.comnet.2022.109453
% This paper confirms and quantifies the four factors, adds a fifth: upload rates (a refinement of bandwidth)

\section{DATA TO DECISION WORKFLOW}
%TODO: desired, not CURRENTLY possible
\begin{figure*}
    \centering
    \includegraphics[width=\textwidth]{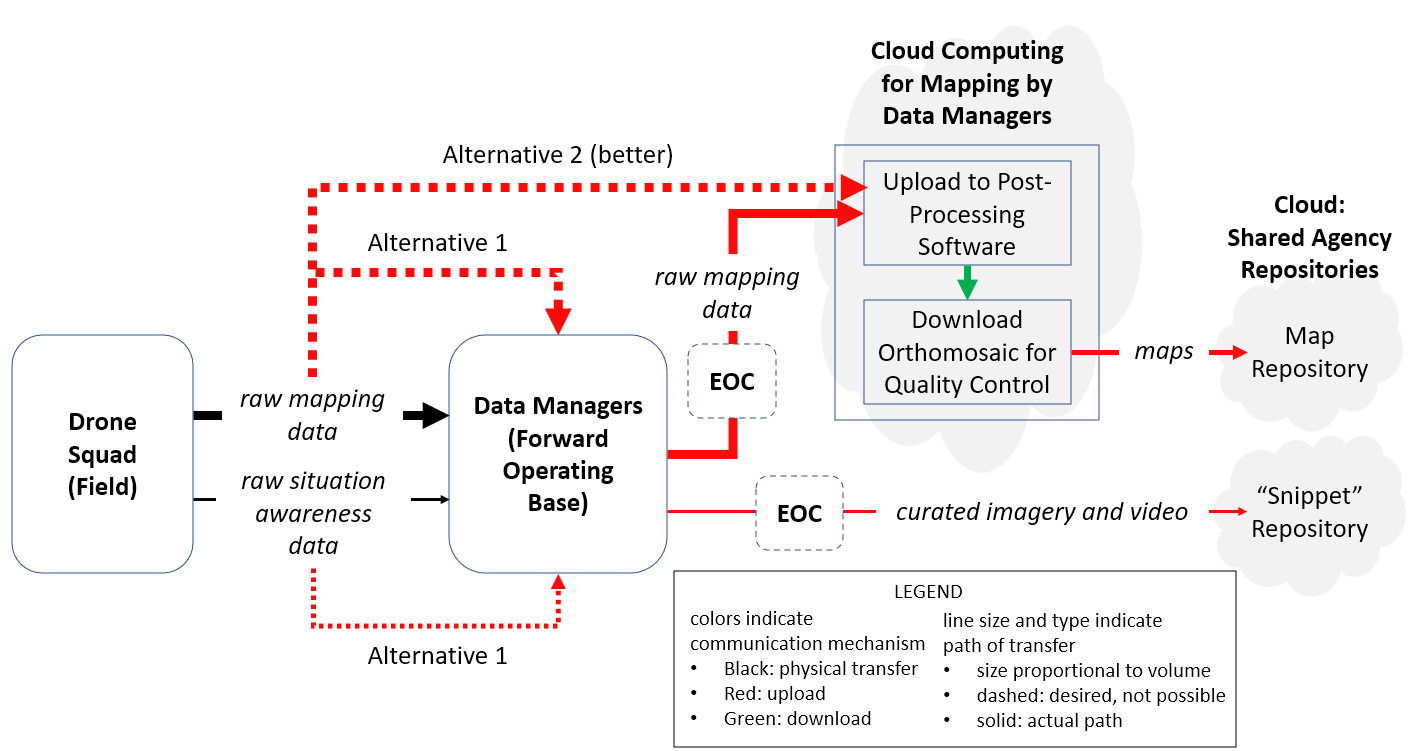}
    \caption{Drone data-to-decision workflow}
    \label{fig:data2decision}
\end{figure*}

A novel attribute of the Hurricane Ian analysis is that it considers the network communications  for the entire drone data-to-decision workflow. Fig.~\ref{fig:data2decision} provides an overview of the drone data-to-decision workflow at Hurricane Ian that is consistent with that observed by the authors at Harvey, Irma, Michael, and Ida. \textit{Drone squads} rally near dawn at the Forward Operating Base (FOB), see Fig.~\ref{fig:map}, receive mission tasking, and deploy to locations throughout the affected area (shaded red area on Fig.~\ref{fig:map}). A squad may be expected to drive or be transported by helicopter to multiple locations before returning. After each drone mission, which may take multiple flights or sorties, the squad will remove from the drone's SD card either \textit{raw mapping data} of damage (a data set of images taken over an area of interest) or \textit{raw situation awareness data} for rapid infrastructure and damage assessment (illustrative videos and still images). The raw mapping data requires post-processing to create a map. A high-resolution map from the same raw imagery requires uploading to Cloud for processing on high performance computers, while a low-resolution map can be rapidly generated on a laptop in under an hour. Generally, given the operations tempo and logistics of missions, drone squads do not have the time or software to post-process maps or curate the most informative imagery and video snippets, rather they flood responders with large volumes of unfiltered data. After the last mission is completed, the squad brings the data to the \textit{data managers} at the FOB to handle that process. The data managers complete the post-mission actions and then push the final dataset to \textit{shared agency repositories}, allowing local, state, and federal stakeholders access.

The red lines in Fig.~\ref{fig:data2decision} indicate wireless communication paths, with solid red indicating actual attempted transmission and dashed red showing ideal paths. Ideally, squads would be able to upload the raw data from a mission as soon as the drone has landed and the SD card is extracted, either to the FOB or directly to the Cloud. The squads would be uploading while moving to the next mission location or returning to the FOB. Typically, the majority of squads would return in the late afternoon, nearly simultaneously, physically transporting the entire day's worth of data. The data managers would then upload the mapping data to the Cloud, monitor it to make sure the upload was not interrupted, download data products and perform quality control checks, then push curated data products to agencies over wireless. As will be seen in the remainder of this paper, none of the paths in red, either solid or dashed, were practical because the wireless infrastructure was not available in the field and did not have sufficient throughput. As a result, the data managers had to physically transport the data products to the Collier County Emergency Operations Center, the nearest facility with an intact high speed wired network. 

\section{DATA COLLECTION}

In the response to Hurricane Ian, FL-UAS1 squads performed two main types of missions resulting in distinct patterns of data collection.
One type of mission is collection of videos and images for immediate damage and infrastructure inspection.
The second type of mission is collection of images following a pre-programmed, raster type scan over an area of interest in order to construct an orthographic map of the area from post-processing software.
The data collection process differed each day of the response due to a number of factors including: physical access to affected areas, operational objectives, and availability of personnel.
The data collection operations are characterized below.
%TODO: Should we cite to a table of the GB/files of data collected. Like what we have in fig:extensions
\textit{September 29, 2022}.
This was the first day of the response and data collection. 
On this date the response was primarily interested in determining the integrity of various large pieces of infrastructure, such as bridges and major roads. 
The data collection consisted of teams being sent to areas of suspected damage to collect opportunistic videos of this infrastructure.
As a result the data collected on this date consisted of a lesser number of very large videos, and occasional photos. In total four squads collected 77 files totalling 46GB.

\textit{September 30, 2022}.
The second day of the response represented a continuation of the previous day but transitioned from imagery of large infrastructure to that of the damage to residential and commercial centers.
This day also contained a mission to central Florida to assess inland flooding.
The data collected on this day was marked primarily by several large video files.
However, unlike the previous day, this day contained FL-UAS1's first mapping mission to Fort Myers Beach.
This resulted in a similar volume of data to the first day: five squads collected 47GB, but consisting of 938 files.
It was also on this day that limitations with the wireless networks that were available around the FOB were observed.
As a result opportunistic collection of network speeds began.

\begin{figure}
 \centering
 \includegraphics[width=\linewidth]{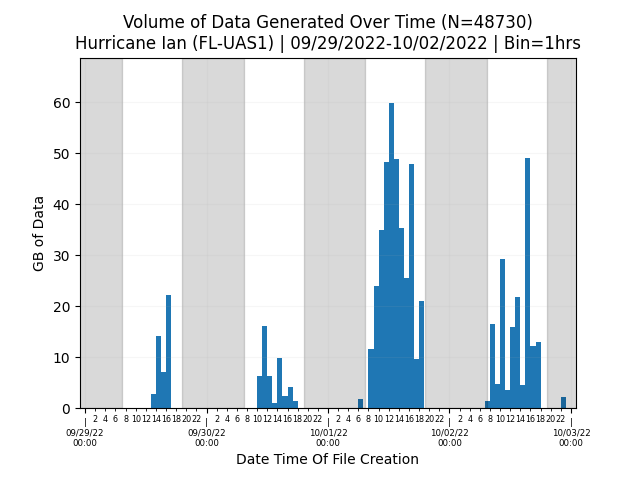}
 \caption{Data generated by FL-UAS1 during the response to Hurricane Ian. Shading indicates time from sunset to sunrise.}
 \label{fig:data_collection}
\end{figure}

\textit{October 1, 2022}.
%The third day represented a significant change from the prior two days.
On this day four more squads joined the response resulting in substantially more capacity. 
This day also marked a change in the type of mission to which squads were dispatched. 
Unlike the previous two days which focused heavily on the opportunistic collection of video imagery, this day transitioned to missions focused on mapping and damage assessment in relative equal proportion. Nine squads collected 33000 files totalling 369GB and opportunistic collection of network speeds continued.

\textit{October 2, 2022}.
Throughout the fourth day of the response, the data collection efforts focused almost exclusively on the collection of overhead still imagery for the purposes of constructing maps.
However, on this date, squads were being sent to mission locations that were further away from the location of the FOB, resulting in an increase in transit time, and thus less time spent collecting data. 
As a result, almost half as much data and files were collected as the previous day.
Nine squads collected 14715 files totalling 174GB and the opportunistic observations of network speeds continued.

\section{FINDINGS}

 In this section, several findings are discussed that are of relevance to the research community and those developing wireless infrastructure for use in large scale disasters. 
\begin{figure}
    \begin{subfigure}{\columnwidth}
    \centering
        \includegraphics[width=9cm]{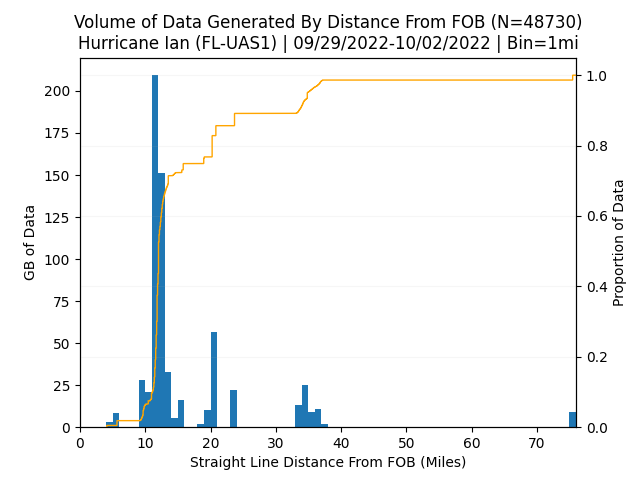}
    \end{subfigure} 
    %\bigskip

            \begin{table}[H]
            \centering
            \setlength\tabcolsep{4pt}
            \begin{tabular}{|l|llllll|}
\hline
Distance & \multicolumn{1}{l|}{p25} & \multicolumn{1}{l|}{p50} & \multicolumn{1}{l|}{p75} & \multicolumn{1}{l|}{p90} & \multicolumn{1}{l|}{p95} & p99  \\ \hline
File Count    & 11.6{\tiny mi}                     & 11.9{\tiny mi}                     & 12.5{\tiny mi}                     & 15.8{\tiny mi}                     & 34.5{\tiny mi}                     & 36.7{\tiny mi} \\
Data Volume   & 11.6{\tiny mi}                     & 12.0{\tiny mi}                     & 18.9{\tiny mi}                     & 33.5{\tiny mi}                     & 34.8{\tiny mi}                     & 75.4{\tiny mi} \\ \hline
\end{tabular}
        \end{table}
    \caption{Data collected in GB versus the distance from the FOB. Percentiles are included for reference.}
    \label{fig:data2fob}
\end{figure}

 One finding of interest is that the general area of operations extended as far as 70 miles from the FOB. 
 This data is discussed in Fig.~\ref{fig:data2fob}.
 To summarize this figure: 50\% of all mission data was collected approximately 12 miles away from the FOB and more than 90\% of all mission data was collected more than 11 miles away from the FOB.
 In fact, 75\% of all mission data was collected between approximately 10 and 13 miles away from the FOB.
 This presents us with narrow range where the vast majority of data is collected, however, it should be noted that some missions extend as far as 70 miles straight line distance from the FOB.

\begin{figure}
    \centering
    \begin{subfigure}{\columnwidth}
    \includegraphics[width=\linewidth]{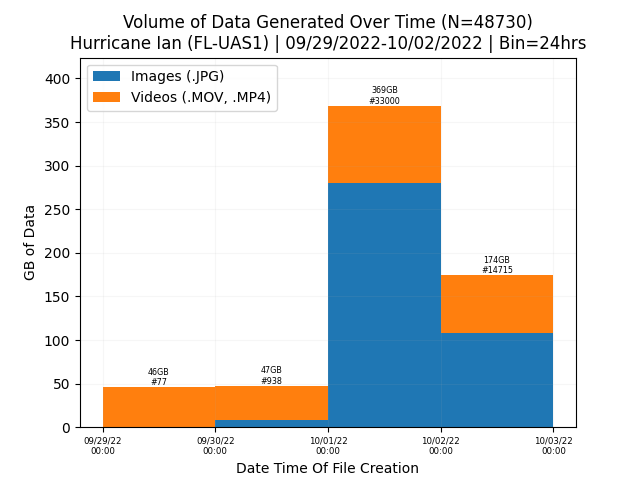}
    \end{subfigure}

    \begin{table}[H]
    \centering
    \setlength\tabcolsep{4pt}
\begin{tabular}{|l|llllll|}
\hline
File Size & \multicolumn{1}{l|}{p25} & \multicolumn{1}{l|}{p50} & \multicolumn{1}{l|}{p75} & \multicolumn{1}{l|}{p90} & \multicolumn{1}{l|}{p95} & p99    \\ \hline
Images {\tiny (N=48592)}    & 6.5{\tiny MB}                      & 8.8{\tiny MB}                      & 10.4{\tiny MB}                     & 11.5{\tiny MB}                     & 12.3{\tiny MB}                     & 14.5{\tiny MB}   \\
Videos {\tiny (N=190)}     & 204.1{\tiny MB}                    & 1.0{\tiny GB}                   & 2.2{\tiny GB}                   & 3.3{\tiny GB}                   & 3.6{\tiny GB}                   & 4.0{\tiny GB} \\ \hline
\end{tabular}
\end{table}

    \caption{Volume of data collected per day by file types. 
    %Images are shown in blue and videos are shown in orange.
    A clear shift can be seen in the distribution along the 72 hour mark. Included for reference are the percentiles of the distribution of file sizes for both videos and images.}
    \label{fig:extensions}
\end{figure}

Next, there is an obvious change in the distribution of data between the first 72 hours and following time periods.
This is reflected in Fig.~\ref{fig:extensions}. In the first two days, the majority of data (in terms of GB) was made up of video files, and comparably small amounts of photos. 
However, along with the start of the third day of data collections, and as FL-UAS1 approached transition 72 hours of operation, the data collection changes to being proportionally many more images and fewer videos, both in terms of file count and GB. 
This imposes a need on the wireless network to handle a lesser number of large files in the initial response, and a larger number of comparatively smaller files after the first 72 hours. In fact less than 25\% of data was collected in the first 48 hours.

This change in data was also marked by the increased capacity of FL-UAS1 as additional personnel joined the response.
On the first day, the response started with four squads, becoming five on the second day, and reaching nine on the third and fourth days.
This increase in personnel demonstrates how the distribution of data collection changes, but also highlights the need for a wireless network that can handle an increasing number of endpoints as the additional resources are added to the response.

\begin{figure}
    \centering
    \includegraphics[width=\linewidth]{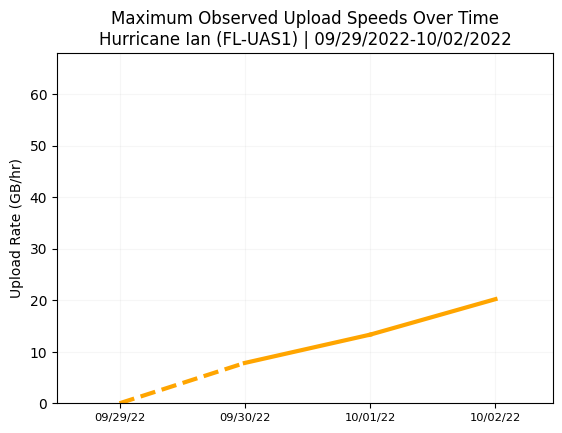}
    \caption{Maximum observed upload rate at the FOB at each date in GB/hr. On 9/29 no observations of wireless upload speeds were made. The upload rate on that date has been approximated as 0 GB/hr.}
    \label{fig:upload_rates}
\end{figure}

Another finding is that network speeds appear to slowly recover in the field each day. %TODO: Do we need to explain this point further, should we provide a hypothesis as to why this is the case
This is shown by the gold line in Fig.~\ref{fig:upload_rates} and Fig.~\ref{fig:informationCurve}. 
Although the line is linear, it is overall consistent with the projected empirical availability (blue).
As expected, the slow recovery is at odds with the timing of data products as per \cite{ccc:EmergencyInformatics}. 
Based on the guidance in the literature, this slow recovery of upload rates is insufficient to support the operational needs of a team such as FL-UAS1.

\begin{figure*}
    \centering
    %TODO: INCLUDE GB/hr
    \includegraphics[width=13.9cm]{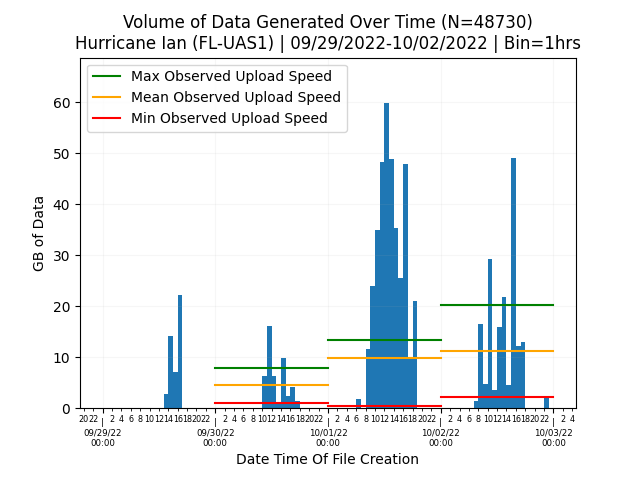}
    \caption{Data collected by FL-UAS1 compared to the upload speeds observed in the field. Horizontal bars represent upload speeds in GB/hr. Any time a blue histogram bin surpasses a colored horizontal line, more data was collected than could have been uploaded. %This plot highlights how, every day, FL-UAS1 experienced at least one hour where the volume of data collected exceeded the best case capacity of the wireless networks available.
    }
    \label{fig:bandwith_saturation}
\end{figure*}

This lack of sufficient wireless connectivity, which has been predicted by past work in the literature, is seen empirically in the field with FL-UAS1.
In Fig.~\ref{fig:bandwith_saturation} the volume of data collected in GB upload rates that were observed in the field are compared.
The figure shows that FL-UAS1 frequently, and at least once per day, collected data at a rate that exceeded even the best case upload speed that was observed in the field.
This presented a scenario where there was more to upload than the network could handle. 
As a result of this FL-UAS1 had to choose which data to upload and when, leading to and lack of information flow to decision makers and resulting response friction.

%TOM:Not sure if this figure adds much at this point considering how the majority of the conclusions will be arrived at in fig:bandwidth_saturation

\section{SUMMARY OF FINDINGS}

These findings highlight several shortfalls of the current wireless infrastructure. These shortfalls are discussed below below along with development considerations for future wireless networks.

\begin{enumerate}[leftmargin=0.5cm]

\item The data-to-decision workflow assumes high upload rates of connectivity which can support burstiness; high download rates are not important.
Burstiness stems from sUAS storing high resolution data onboard, recovered only when the drone has landed (20-40 minutes) and from teams in the field conducting  multiple missions over 6-12 hours before returning to the FOB. 

\item The teams currently do not have sufficient connectivity in the field to a) live stream, b) upload bursts of tactical data directly to agencies, or c) upload those mission bursts for cloud computing.
The teams instead must transfer data when they return to the FOB at the end of the day, leading to bursts of data, which is exacerbated by multiple teams returning at one time.

\item The FOB itself does not have sufficient connectivity to handle the burst of data  from multiple teams arriving at the end of the day; data managers need to plan to physically transfer data to the nearest EOC.
The data generated exceeded the observed data rates of cellular (ATT, Verizon, FirstNet, PSEC) and satellite providers (commercial and personal). A wireless connectivity “bubble” around the FOB would have to extend on the order of 10-40 miles (90\% of data is collected $>$10mi away), consistent with Harvey, Michael, and Irma.

\item The greatest need for a large-scale wireless network at the FOB is in the first 3 days (also corresponding to the 72 hour default response phase).
If the network is not available by the end of the second day when missions shift to mapping, the latency between data collection and distribution to decision makers will become moot.

\item The number of teams and associated data demands dynamically change.
Mission tasking may increase unpredictably and more teams may join the response. 
This would increase the need for team-owned connections in field and burstiness at the FOB. The average squad generated 1874 files totalling 24GB per day.
\end{enumerate}

\section{CONCLUSIONS}
%RRM

The analysis of Hurricane Ian provides a quantitative analysis of wireless demands and current gaps for supporting the drone data-to-decision workflow in a disaster. The results document the temporal aspects of wireless availability, and that the first 72 hours are the most critical for impact of information on the decision-making process, yet there is not sufficient  bandwidth with either FirstNet, satellite, or consumer cellular service. An effective wireless network would need to provide  communications for mobile drone squads working from 10 to 40 miles from the FOB (spatial distribution). 
The gap in bandwidth is upload rate, as the primary data transfers are uploading from drone to either data manager, agency, or the cloud; and from data manager to either the cloud or agencies. Downloading occurs rarely. 
The lack of high upload rates is exacerbated by burstiness, where drone squads attempt to upload mission data after the drone lands or from all missions at the end of the day. 

The results suggest that future designs of drones, software, and networks should focus on advancing edge computing, given 
the lack of high bandwidth wireless communications during the first 72 hours.
Drone systems and software should be adaptable to whatever communications infrastructure is available (Fog computing), with physically transporting data as a likely occurrence. Although outside of the scope of this paper, the results also suggest that drone control schemes should not expect to rely on internet during a disaster. 
The results also highlight that the drone data-to-decision workflow network demands are notably different from normal consumer streaming media instances. Disaster response requires advances in  upload, not download, rates  over a communications infrastructure that is immediately available. 
%
% Wireless and robotics experts need to work together to create better a more suitable Cloud/Edge partitioning for disasters.Technologists and emergency managers need to work together to identify high impact data post processing steps that can be moved to edge computing in order to minimize dependence on 

%%%%%%%%%%%%%%%%%%%%%%%%%%%%%%%%%%%%%%%%%%%%%%%%%%%%%%%%%%%%%%%%%%%%%%%%%%%%%%%%

\section*{ACKNOWLEDGMENT}

This material is based upon work supported by the National Science Foundation under Grant No.CMMI-2306453. The authors thank Dr. Walt Magnussen and the Internet 2 Technology Evaluation Center for their support and loan of FirstNet devices.

%%%%%%%%%%%%%%%%%%%%%%%%%%%%%%%%%%%%%%%%%%%%%%%%%%%%%%%%%%%%%%%%%%%%%%%%%%%%%%%%

%\begin{thebibliography}{99}
\bibliographystyle{IEEEtran}
\bibliography{root}
%\end{thebibliography}

\end{document}